\documentclass[11pt]{article}

\usepackage[]{acl}

\usepackage{times}
\usepackage{latexsym}

\usepackage[T1]{fontenc}

\usepackage[utf8]{inputenc}

\usepackage{microtype}

\usepackage{inconsolata}

\usepackage{bm}
\usepackage{array}
\usepackage{xspace}
\usepackage{mathrsfs}
\usepackage{amssymb}
\usepackage{amsmath}
\usepackage{enumitem}
\usepackage{booktabs}
\usepackage{graphicx}
\usepackage{multirow}
\usepackage{multicol}
\usepackage{makecell}
\usepackage{listings}
\usepackage{subfigure}
\usepackage{tablefootnote}
\usepackage[normalem]{ulem}
\usepackage{algorithm}
\usepackage{algpseudocode}

\usepackage{nicematrix}
\usepackage{booktabs}
\usepackage[most]{tcolorbox}
\usepackage{color,colortbl}
\definecolor{color1}{HTML}{006EB8}
\definecolor{color2}{HTML}{009B55}
\definecolor{color3}{HTML}{00A99A}
\definecolor{color4}{HTML}{3C8031}
\definecolor{color5}{HTML}{006795}
\definecolor{color6}{HTML}{00AEB3}

\setlist[itemize]{leftmargin=5mm, itemsep=0mm}

\definecolor{comment}{rgb}{0,0.6,0}
\definecolor{key}{rgb}{0.84, 0.44, 0.84}
\definecolor{func}{rgb}{0.16, 0.72, 0.86}
\definecolor{var}{rgb}{1, 0.33, 0.33}
\definecolor{true}{rgb}{1, 0.78, 0.02}

\newcommand{\method}{PCPO\xspace}
\newcommand{\ie}{\emph{i.e.,}\xspace}
\newcommand{\eg}{\emph{e.g.,}\xspace}
\newcommand{\etc}{\emph{etc.}\xspace}

\DeclareMathOperator*{\argmax}{arg\,max}

\title{Probability-Consistent Preference Optimization for \\ Enhanced LLM Reasoning}

\author{%
    Yunqiao Yang$^{1,5}$\thanks{Equal contribution.} \quad Houxing Ren$^{1}$\footnotemark[1] \quad Zimu Lu$^{1}$ \quad Ke Wang$^{1}$ \quad Weikang Shi$^{1}$ \\
    \textbf{Aojun Zhou}$^{1}$ \quad \textbf{Junting Pan}$^{1,3}$ \quad \textbf{Mingjie Zhan}$^{2}$\footnotemark[2] \quad \textbf{Hongsheng Li}$^{1,3,4}$ \thanks{Corresponding author.} \\
    $^{1}$CUHK MMLab, $^{2}$SenseTime Research \\ $^{3}$CPII under InnoHK, $^{4}$Shanghai AI Laboratory, $^{5}$Zhiyuan College, SJTU  \\
    \texttt{yangyunqiao7@gmail.com} \quad \texttt{zhanmingjie@sensetime.com} \quad \texttt{hsli@ee.cuhk.edu.hk}
}

\begin{document}
\maketitle
\begin{abstract}
Recent advances in preference optimization have demonstrated significant potential for improving mathematical reasoning capabilities in large language models (LLMs). While current approaches leverage high-quality pairwise preference data through outcome-based criteria like answer correctness or consistency, they fundamentally neglect the internal logical coherence of responses.  To overcome this, we propose Probability-Consistent Preference Optimization (PCPO), a novel framework that establishes dual quantitative metrics for preference selection: (1) surface-level answer correctness and (2) intrinsic token-level probability consistency across responses. Extensive experiments show that our PCPO consistently outperforms existing outcome-only criterion approaches across a diverse range of LLMs and benchmarks. Our code is publicly available at \url{https://github.com/YunqiaoYang/PCPO}.
\end{abstract}

\section{Introduction} \label{sec: intro}

In recent years, enhancing the mathematical reasoning ability of Large Language Models~\cite{GPT42023ABS230308774, Palm2023ABS230510403, LLama2023ABS230213971, LLama22023ABS230709288, Qwen2023ABS230916609, Mistral2023ABS231006825, Mixtral2023ABS240104088, Claude2023Anthropic, Qwen252024Yang} (LLMs) has emerged as an important research direction~\cite{ahn2024large,minaee2024large}. Among various approaches, Direct Optimization Preference (DPO)~\cite{rafailov2024direct} is widely used due to its simplicity and efficiency. 
Since its introduction, numerous extensions of DPO have been proposed to further improve mathematical reasoning in diverse ways. For instance, methods such as Self-Rewarding LLMs~\cite{yuan2024self} and iterative DPO~\cite{xu2023some} demonstrate the effectiveness of iterative training strategies. Additionally, constructing high-quality pairwise preference data is essential for preference optimization~\cite{bai2022constitutional,yang2023rlcd}. 

To construct high-quality pairwise preference data, previous methods, such as IRPO~\cite{pang2024iterative} and ScPO~\cite{prasad2024self}, select preference training pairs from generated responses that include a Chain-of-Thought~(CoT)~\cite{kojima2022large} process followed by a final answer, have proven particularly effective in advancing mathematical reasoning performance. IRPO~\cite{pang2024iterative} employs gold labels (correct answers) to distinguish between chosen and rejected responses. Specifically, if a response's answer matches the gold label, it is designated as a chosen response; otherwise, it is classified as rejected. On the other hand, ScPO~\cite{prasad2024self} utilizes a voting function to evaluate the self-consistency~\cite{wang2022self} of responses. Responses whose answers appear most frequently are selected as chosen, while those with the least frequent answers are marked as rejected.

\begin{figure*}[ht]
	\includegraphics[width=\linewidth]{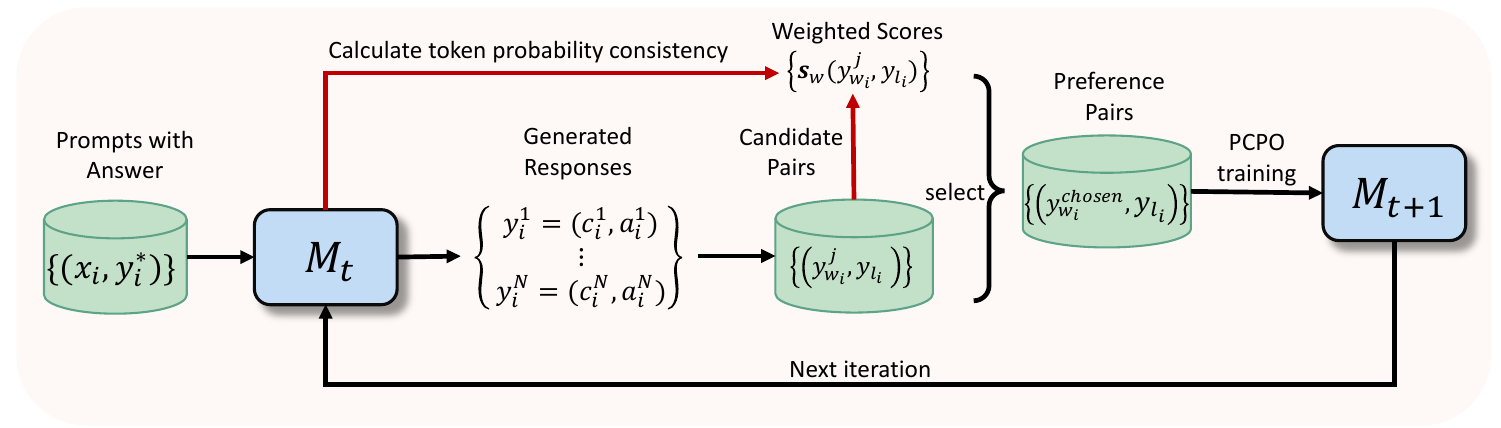}
	\caption{Overview of the \method method. The pipeline mainly consists of three steps. (1) Given a prompt set, utilize $M_t$~($M_0$ as the seed model) to generate responses $y_i^n$ with reasoning $c_i^n$ and answer $a_i^n$, and construct candidate pairs on correctness \S\ref{subsec: candidate}. (2) utilize $M_t$ to calculate weighted scores $s_{w}$ for each pair based on the token probability consistency, and select preference pairs based on it \S\ref{subsec: pair}. (3) train the next iteration model $M_{t+1}$ with the selected preference pairs and \method Loss \S\ref{subsec: loss}.
	}
	\label{fig: method}
	\vspace{-0.5cm}
\end{figure*}

However, both methods focus solely on the correctness or frequency of the final answer while overlooking the internal logical connections or nuanced differences between responses. This limitation restricts the creation of more refined and informative preference training data~\cite{wang2024lrhp}. Consequently, models may have difficulty recognizing subtle yet critical distinctions between chosen and rejected responses during the iterative DPO training process~\cite{furnkranz2010preference,wirth2017survey}.

In this paper, we propose a novel method called \textbf{P}robability-\textbf{C}onsistent \textbf{P}reference \textbf{O}ptimization (\method), which leverages both the final answer and the internal logical connections of responses when selecting preference pairs. Our method is grounded in the principle that the token generation process in LLMs fundamentally involves predicting new tokens based on the highest conditional probability given all existing tokens~\cite{vaswani2017attention,radford2018improving}. Specifically, \method calculates a weighted score between preferred and dispreferred answers by evaluating the conditional probability of each token in the responses~\cite{holtzman2019curious,welleck2020consistency}. Preference pairs are then selected based on the highest weighted scores. This approach provides a quantitative framework for selecting preference pairs by considering not only the correctness of the final answer but also the internal coherence of the responses. By incorporating these factors, \method ensures a more robust and principled selection of preference pairs.

In each iteration, We first use the seed model to generate multiple responses for each math problem, and we construct a candidate preference pair set based on the correctness of the final answer. Then, we calculate the token-level consistency score for all the preference pairs from the candidate pair set. Afterward, we select the preference pairs with the highest token-level weighted score for each problem to construct preference training pairs eventually. Finally, we use the preference pairs selected to train the next iteration model with a modified DPO loss. To validate our method, we apply it to widely used math datasets, including GSM8K~\cite{cobbe2021training}, MATH-500~\cite{hendrycks2021measuring,lightman2023let} Olympiadbench~\cite{he2024olympiadbench} and AMC23~\cite{AMC2023}.To comprehensively demonstrate the effectiveness of our approach, we conduct experiments across a diverse range of seed models, such as Llama-3-8b-Instruct~\cite{dubey2024llama}, Mathstral-7b-v0.1~\cite{jiang2023mistral}, Qwen-2.5-7B-Instruct~\cite{yang2024qwen2} and Qwen-2.5-Math-7B-Instruct~\cite{yang2024qwen2math}. Consistent results across these models showcase the effectiveness of our method.

In summary, our contributions are as follows:

1) We propose Probability-Consistent Preference Optimization (\method), a novel method that leverages both the final answers and the internal connections of the responses to select higher-quality preference pairs for training, thereby enhancing the mathematical reasoning capabilities of seed LLMs.

2) Extensive experiments demonstrate that our method consistently outperforms existing outcome-only criterion approaches (\eg, IRPO, ScPO) across a diverse range of LLMs and benchmarks.

3) Through empirical analysis, we highlight the critical importance of considering the internal connections of responses when selecting preference pairs. This insight paves the way for future research aimed at improving reasoning capabilities through more sophisticated preference pair selection methods.

\section{Method} \label{sec: met}

As depicted in Figure~\ref{fig: method}, our method starts with a pre-trained seed language model and a fixed prompt set of math problems with final answers. The \method pipeline mainly consists of three steps. (1) utilize $M_t$~($M_0$ as the seed model) to generate responses $y_i^n$ with reasoning $c_i^n$ and answer $a_i^n$, and construct candidate pairs on correctness \S\ref{subsec: candidate}. (2) utilize $M_t$ to compute weighted scores $s_{w}$ for each pair based on the token probability, and select preference pairs based on it \S\ref{subsec: pair}. (3) train the next iteration model $M_{t+1}$ with the selected preference pairs and \method loss \S\ref{subsec: loss}. The model will be trained and updated at each iteration, resulting in a series of models $M_1, \dots, M_T$.

\subsection{Construct Candidate Pairs} \label{subsec: candidate}

We assume we have an initial model $M_0$, and a prompt set $D = \{(x_i, y_i^*)\}$ containing questions $x_i$ and their correct answers $y_i^*$.
We focus on the process applied to a specific prompt $x_i$, and therefore, we omit the subscript $i$ for simplicity in the following subsections.

\paragraph{Response Generation.}  In each iteration, we first use the current model $M_t$ to generate N different responses for the prompt $x$, \ie $Y = \{y^n = (c^n,a^n) \sim M_t(x)\}$  and $n \in [N]$, where $c^n,a^n$ represents the Chain-of-Thought reasoning steps and the prediction answer. If the prediction answer $a^n$ of the response $y^n$ equals the gold answer $y^*$, we put the response into a response subset $Y_w$, otherwise $Y_l$.

\paragraph{Prepare candidate pair set.}
Assume that there are $p$ chosen responses $y_{w}$ in $Y_w$ and $q$ rejected responses $y_{l}$ in $Y_l$, where $p$ and $q$ satisfy the condition $p + q = N$. To generate all possible preference pairs, we use the cartesian product~\cite{hewitt1955symmetric} between the set $Y_w$ and  $Y_l$ to construct \( p \times q \) pairs, denoted as $Y_w \times Y_l = \{(y_w, y_l) \mid y_w \in Y_w, y_l \in Y_l\} $. However, due to computational constraints, we must limit the number of candidate pairs. To achieve this, we employ the Levenshtein distance technique \cite{heeringa2004measuring} to filter the candidate pairs effectively. Its rationale is discussed in Appendix~\ref{app: leven}. 

The Levenshtein distance measures the minimum number of edits required to transform one sequence into another, serving as a metric for sequence similarity. For each rejected response \( y_{l} \), we compute its Levenshtein distance with all chosen responses \( \{y_{w}\} \) and select the top \( k \) pairs with the smallest distances as candidate pairs. This process results in the candidate pair set \( \mathcal{C}_t^{pairs} = \{(y_{w}^{j}, y_{l})\} \), where $j=1,2,...,\min(p,k)$ represents the number of candidate chosen responses for each rejected response $y_l$.
\subsection{Construct Preference Pairs} \label{subsec: pair}

In this step, we first introduce the concept of \textit{token probability consistency} ($c_t$), a token-level metric derived from the standard cross-entropy formulation for individual tokens~\cite{vaswani2017attention,radford2018improving,hong2024orpo}:
\begin{equation}
	\mathcal{L}_t = -\log P(x_i|x_{<i}).
\end{equation}
Next, we define the \textit{pair-weighted score}($s_w$), a pair-level metric computed from the \textit{token probability consistency} values of the chosen and rejected responses within a pair. Based on this score, we selectively extract preference training pairs from the candidate pairs set $\mathcal{C}_t^{pairs}$.

\paragraph{Calculate token probability consistency.} \label{subsec: consistency}
For each response $y$ in a certain candidate pair $(y_{w},y_{l})$, we perform the following steps. First we tokenize $y$ into a sequence $\{y^t\}$ using the tokenizer of the current iteration model $M_t$, where $t=1,2,...,l$ and $l$ denotes the length of the token sequence $\{y^t\}$. We then infer $M_t$ to obtain the casual conditional probability $P_{M_t}(y^t|y^{<t},x)$ for each token. 

With the tokenized pairs and their corresponding token probabilities, we proceed to the next step: employing a matching function $\mathcal{M}$ in Appendix~\ref{app: match} to align the common tokens between the two responses in a pair sequentially. This allows us to compute the \textit{token consistency score}($c_t$), which is defined as
\begin{multline}
	c_t\left( y_w|y_l \right) = \exp ( - |\log  P_w\left( y_t|x,y_{<t} \right) \\
	-\log  P_l\left( y_t|x,y_{<t} \right)|) \,\,,y_t\in \mathcal{M}(y_w, y_l).
\end{multline}

The concept of comparing token-level losses draws inspiration from recent works such as~\citet{christopoulou2024sparsepo}, which emphasizes sparse token-level optimization, and~\citet{lin2024critical}, which highlights the importance of critical tokens in alignment tasks. Additionally, the use of exponential mapping aligns with the design principles of ORPO's odd-one-out loss~\cite{hong2024orpo}, as both approaches aim to transform token-level differences into probabilistic metrics for more effective optimization. This combination of ideas provides a principled foundation for our token probability consistency framework.

\paragraph{Calculating pair-weighted score.} 
The token-level consistency score \( c_t \) is a normalized metric ranging between 0 and 1, where a higher value indicates a smaller difference in logarithmic probabilities between the chosen and rejected responses for a given token. Since the logarithmic probability represents a conditional probability, a higher score suggests that the preceding tokens provide the most relevant context for predicting the current token~\cite{vaswani2017attention,radford2018improving}. To compute the overall score \( s \), we aggregate the token-level consistency scores \( c_t \) across all matched tokens. However, since the number of matched tokens varies with the length of the responses, we normalize the final score by dividing it by the total length of the preference pair. This yields the pair-weighted score \( s_{w} \) for each preference pair, defined as
\begin{equation}
	s_{w}\left( y_{w} | y_{l} \right) = \frac{\sum_t c_t\left(y_{w} | y_{l} \right)}{l_{y_{l}}},
	\label{eq: score}
\end{equation}
where \( l_{y_{l}} \) denotes the length of the token sequence in the rejected response. This normalization ensures that the score is robust to variations in response length and provides a fair comparison across preference pairs.

\paragraph{Select preference pairs.}  
Given that we have performed the process in Section~\ref{subsec: consistency} for all the candidate pairs $(y_{w}^j, y_{l})$. Next, we select the preference pair with the highest pair-weighted score for the rejected responses \( y_{l} \). Specifically, for a given rejected response \( y_{l} \), we choose the corresponding chosen response \( y_{w}^{chosen} \) that maximizes the pair-weighted score \( s_w \). This ensures that the chosen response exhibits the strongest token-level consistency and correlation with the rejected response, making it the most suitable candidate for preference optimization~\cite{holtzman2019curious,welleck2020consistency}. The resulting set of selected preference pairs can be formally represented as
\begin{align}
	&\mathcal{S}_t^{pairs} = \{( y_{w}^{chosen}, y_{l})\} \\
	&= \left\{\argmax_{(y_{w}^j y_{l})} s_w\left(y_{w}^j | y_{l}\right) \,\bigg|\, y_{w}^j, y_{l} \in \{(y_{w}^j, y_{l})\} \right\},
	\label{eq: pair}
\end{align}
where \( \mathcal{S}_t^{pairs} \) denotes the final set of selected preference pairs for the prompt x, and \( \argmax \) identifies the chosen response \( y_{w}^{chosen} \) that maximizes the pair-weighted score \( s_w \) for a given rejected response \( y_{l} \). This selection process ensures that the chosen pairs are optimized for token-level consistency and alignment with human preferences, while maintaining a strong correlation between the chosen and rejected responses.

\subsection{\method Loss Function}\label{subsec: loss}
We design our \method loss function as follows:
\resizebox{\linewidth}{!}{
	\begin{minipage}{\linewidth}
		\begin{align*}
			&\mathcal{L}_{\method}(y^+, y^- | x)\!
			=\! \\
			&\underbrace{- s_w(x) \log \sigma\left(\beta \log \frac{M_\theta(y^+\!\mid\!x)}{M_t(y^+ \mid x)} - \beta \log \frac{M_\theta( y^-\!\mid\!x)}{M_t(y^-\!\mid x)}\right)}_{\text{Weighted DPO Loss}} \\
			&\underbrace{-\frac{\alpha s_w(x)}{|y^+|}\log M_\theta(y^+\!\mid\!x)}_{\text{Weighted NLL Loss}}.
			\label{eq: loss}
		\end{align*}
		\vspace{0.05em}
	\end{minipage}
}

The loss function integrates a pair-weighted score \( s_w \) into both DPO and NLL losses, inspired by IRPO~\cite{pang2024iterative} and ScPO~\cite{prasad2024self}. The weighted DPO loss and the weighted NLL loss, dynamically prioritize pairs with high token-level consistency, akin to sparse alignment strategies in SparsePO~\cite{christopoulou2024sparsepo}. It also adaptively balances language modeling with preference alignment, similar to  ScPO's self-consistency weighting.

The use of \( s_w \) as a dynamic weighting mechanism is grounded in token-level consistency principles from ~\citet{zeng2024token} and  ~\citet{lin2024critical}, while the inclusion of NLL loss ensures stable optimization, as highlighted in IRPO~\cite{pang2024iterative}. This design enables adaptive sample weighting, robustness to sequence length variations, and flexible optimization through parameters \( \beta \) and \( \alpha \). The pair-weighted score \( s_w \) serves as a key innovation, enhancing the training process's effectiveness and interpretability.

\section{Experiment setup} \label{sec: setup}

\begin{table*}[ht]
	\small
	\centering
	\setlength{\tabcolsep}{7.0pt}
	\begin{NiceTabular}{l|cc|cc|cc|cc}
		\toprule
		{\bf Metric} & \multicolumn{2}{c}{\bf GSM8K} & \multicolumn{2}{c}{\bf MATH-500} & \multicolumn{2}{c}{\bf Olympiadbench} & \multicolumn{2}{c}{\bf AMC23}\\
		~~Iteration & Pass@1 & Maj@8 & Pass@1 & Maj@8 & Pass@1 & Maj@8 & Pass@1 & Maj@8 \\
		\midrule
		\rowcolor{white!7}
		\rowcolor{white!7} {\small \hspace{-2mm} \emph{Llama3-8B-Instruct} } &  & & &  &  &  & &\\
		\rowcolor{white!7} ~~Seed $M_0$ & 71.3 & 81.6 & 30.8 & 34.2 & ~~8.1 & 10.2 & 10.0 & ~~7.5\\ %
		\rowcolor{white!7} ~~IRPO $M_1$ & 79.1 & 86.4 & 29.4 & 35.6 & ~~7.3 & 10.0 & ~~~~0 & 17.5\\ %
		\rowcolor{white!7} ~~IRPO $M_2$ & 81.1 & 88.4 & 30.6 & 36.6 & ~~6.7 & ~~9.8 & ~~~~0 & 12.5\\ %
		\rowcolor{white!7} ~~ScPO $M_1$ & 79.3 & 87.5 & 30.2 & 34.6 & ~~6.4 & ~~10.4 & ~~7.5 & 15.0\\ %
		\rowcolor{white!7} ~~ScPO $M_2$ & 81.6 & 88.6 & 32.2 & 36.4 & ~~7.9 & ~~10.5 & ~~5.0 & 17.5\\ %
		\rowcolor{gray!7} ~~\method (ours) $M_1$ & 80.1 & 87.8 & 32.2& 36.6 & ~~7.9& ~~9.5 & \textbf{15.0} & \textbf{22.5} \\ %
		\rowcolor{gray!7} ~~\method (ours) $M_2$ & \textbf{82.8} & \textbf{88.9} & \textbf{33.2}& \textbf{38.4} & ~~\textbf{9.5}& \textbf{11.7} & 10.0 & 20.0 \\ %
		\midrule
		\rowcolor{white!7}
		\rowcolor{white!7} {\small \hspace{-2mm} \emph{Mathstral-7B-v0.1} } &  & & &  &  &  & &\\
		\rowcolor{white!7} ~~Seed $M_0$ & 84.3 & 91.4 & 57.2 & 63.2 & 21.8 & 26.7 & 25.0 & 40.0\\ %
		\rowcolor{white!7} ~~IRPO $M_1$ & 87.0 & 92.3 & 57.2 & 63.4 & 23.6 & 29.0 & 20.0 & 32.5\\ %
		\rowcolor{white!7} ~~IRPO $M_2$ & 87.7 & 91.4 & 58.4 & 66.8 & 24.6 & 29.2 & 20.0 & 30.0\\ %
		\rowcolor{white!7} ~~ScPO $M_1$ & 87.1 & 92.0 & 57.4 & 65.4 & 23.4 & 30.5 & 22.5 & 27.5\\ %
		\rowcolor{white!7} ~~ScPO $M_2$ & 87.6 & 92.3 & 60.4 & 66.8 & 24.1 & 30.7 & 27.5 & 40.0\\ %
		\rowcolor{gray!7} ~~\method (ours) $M_1$ & 87.9 & 91.9 & 58.6& 66.4& 24.9 & 29.2 & 20.0 & 37.5 \\ %
		\rowcolor{gray!7} ~~\method (ours) $M_2$ & \textbf{89.0} & \textbf{92.3} &\textbf{61.8} &\textbf{69.4} & \textbf{25.2} & \textbf{32.1} & \textbf{32.5} &\textbf{47.5} \\ %
		\midrule
		\rowcolor{white!7}
		\rowcolor{white!7} {\small \hspace{-2mm} \emph{Qwen2.5-7B-Instruct} } &  & & &  &  &  & &\\
		\rowcolor{white!7} ~~Seed $M_0$ & 92.3 & 94.0 & 76.4 & 81.2 & 38.5 & 44.9 & 47.5 & 60.0\\ %
		\rowcolor{white!7} ~~IRPO $M_1$ & 92.2 & 93.9 & 75.2 & 80.4 & 37.9 & 43.3 & 50.0 & 55.0\\ %
		\rowcolor{white!7} ~~IRPO $M_2$ & 92.3 & 93.9 & 77.6 & 81.2 & 40.1 & 45.0 & 52.5 & 57.5\\ %
		\rowcolor{white!7} ~~ScPO $M_1$ & 92.2 & 94.1 & 76.8 & 80.8 & 39.9 & 44.4 & 55.0 & 60.0\\ %
		\rowcolor{white!7} ~~ScPO $M_2$ & 92.3 & 93.9 & 76.8 & 81.4 & 39.9 & 44.7 & \textbf{57.5} & 60.0\\ %
		\rowcolor{gray!7} ~~\method (ours) $M_1$ & 92.6 & \textbf{94.5} & 76.4& 81.8 & 39.9& \textbf{45.9} & 45.0 & 62.5 \\ %
		\rowcolor{gray!7} ~~\method (ours) $M_2$ & \textbf{92.6} & 94.1 & \textbf{78.0} & \textbf{82.4}& \textbf{40.3} & 45.0& \textbf{57.5} & \textbf{65.0} \\ %
		\midrule
		\rowcolor{white!7}
		\rowcolor{white!7} {\small \hspace{-2mm} \emph{Qwen2.5-Math-7B-Instruct} } &  & & &  &  &  & &\\
		\rowcolor{white!7} ~~Seed $M_0$ & 92.9 & 93.9 & 81 & 83.0 & 43.4 & 46.1 & 62.5 & 70.0\\ %
		\rowcolor{white!7} ~~IRPO $M_1$ & 93.1 & 94.0 & 81.2 & 82.8 & 44.1 & 47.4 & 67.5 & 70.0 \\ %
		\rowcolor{white!7} ~~IRPO $M_2$ & 92.7 & 93.9 & 79.8 & 83.6 & 44.6 & 47.7 & 65 & 70.0 \\ %
		\rowcolor{white!7} ~~ScPO $M_1$ & 92.6 & 94.1 & 80.8 & 83.0 & 44.7 & 47.3 & 67.5 & 70.0 \\ %
		\rowcolor{white!7} ~~ScPO $M_2$ & 93.1 & 94.0 & 80.8 & 83.0 & 44.6 & 48.1 & 67.5 & 70.0\\ %
		\rowcolor{gray!7} ~~\method (ours)  & 92.9 & \textbf{94.2} & 80.6 &  83.4 & \textbf{44.9} & 48.7 & \textbf{70.0} & 72.5 \\ %
		\rowcolor{gray!7} ~~\method (ours)  & \textbf{93.3} & 94.1 & \textbf{81.4} & \textbf{83.8} & 44.3 & \textbf{48.7} & 67.5 & \textbf{75.0} \\ %
		\bottomrule
	\end{NiceTabular}
	\caption{ Results of our method \method comparing with the baseline methods on GSM8K, MATH, Olympiadbench, and AMC23. The results are zero-shot Pass@1 and Maj@8 accuracy.}
	\label{tab: math_main}
\end{table*}

\paragraph{Datasets.} We assess the effectiveness of \method across a large and challenging range of mathematical reasoning datasets: 
\textbf{GSM8K} consists of 1.3k high-quality grade school math word problems. 
\textbf{MATH-500} is a curated subset drawn from the MATH dataset comprising 500 challenging competition-style mathematics problems.
\textbf{Olympiadbench} is a test set of mathematics problems from olympiads, designed to assess deep problem-solving skills, creativity, and advanced mathematical reasoning. 
\textbf{AMC23} is a test set of 40 problems from the 2023 American Mathematics Competitions (AMC 12). These problems are renowned for their depth and subtlety, offering a rigorous assessment of reasoning skills and precision.

\paragraph{Metrics.} We report zero-shot Pass@1 and Maj@8 results. The Pass@1 score denotes The greedy decoding accuracy of a single response. The Maj@8 score denotes the accuracy of the majority answer voted from 8 candidate responses~\cite{wang2022self}. More evaluation details are presented in Appendix~\ref{app: eval}.

\paragraph{Training data.} Our training data includes 7.5k GSM8K training set, 7.5k MATH training set, 7.5k subset of Orca-math~\cite{li2024numinamath}, and 7.5k subset of Cn-k12~\cite{li2024numinamath}, 30k in total. In our approach, we don't need to generate new data, and the training data are fixed for all the experiments.

\paragraph{Baselines.} \textbf{Seed Model} uses Chain-of-Thought prompting~\cite{kojima2022large} with greedy decoding, achieving zero-shot Pass@1 and Maj@8 accuracy. 
\textbf{IRPO}~\cite{pang2024iterative}  utilizes iterative training with pairwise preferences at the outcome level, considering the correctness of the final answer when building preference training data. 
\textbf{ScPO}~\cite{prasad2024self} uses an inference-time-only approach that selects the most frequent final answer to build preference training data. Similar to IRPO, ScPO is still an outcome-level method that considers the correctness and the frequency of the final answer.

\paragraph{Implementation details.}
We set $N=16$ to generate responses for the training data, with the temperature of 1 and top-$p=0.95$. For each iteration, we sample 15k training data, training a total of 6 epochs with a useful batch size of 128. We use an initial leaning rate $1.0 \times10^{-7}$ with the cosine scheduler and AdamW optimizer with a warm-ratio of 0.1 for smoother training. The NLL regularization coefficient $\alpha$ is set to 1 and the DPO loss term coefficient $\beta$ is set to 0.5, following ~\citet{prasad2024self}. For the Pass @1 evaluation, we implement greedy decoding with the temperature of 0, and for the Maj@8 evaluation, we set a temperature of 0.95 and top-$p=0.95$. We use one node containing 8 A800 GPUs for training.

\section{Main Rresults} \label{sec: exp}

\begin{table*}[ht]
	\small
	\centering
	\setlength{\tabcolsep}{7.0pt}
	\begin{NiceTabular}{l|cc|cc|cc|cc}
		\toprule
		{\bf Metric} & \multicolumn{2}{c}{\bf GSM8K} & \multicolumn{2}{c}{\bf MATH-500} & \multicolumn{2}{c}{\bf Olympiadbench} & \multicolumn{2}{c}{\bf AMC23}\\
		~~Iteration & Pass@1 & Maj@8 & Pass@1 & Maj@8 & Pass@1 & Maj@8 & Pass@1 & Maj@8 \\
		\midrule
		\rowcolor{white!7}
		\rowcolor{white!7} {\small \hspace{-2mm} \emph{Llama3-8B-Instruct} } &  & & &  &  &  & &\\
		\rowcolor{white!7} ~~Seed $M_0$ & 71.3 & 81.6 & 30.8 & 34.2 & ~~8.1 & 10.2 & 10.0 & ~~7.5\\ %
		\rowcolor{white!7} ~~IRPO+DPO $M_1$ & 79.8 & 88.1 & 29.4 & 34.2 & ~~6.8 & 10.8 & ~~5.0 & 10.0\\ %
		\rowcolor{white!7} ~~IRPO+DPO $M_2$ & 81.7 & 88.2 & 30.0 & 35.8 & ~~7.4 & ~~8.1 & ~~~~0 & ~~5.0\\ %
		\rowcolor{white!7} ~~ScPO+DPO $M_1$ & 79.3 & 86.7 & 29.4 & 37.0 & ~~7.3 & 11.0 & ~~5.0 & 10.0\\ %
		\rowcolor{white!7} ~~ScPO+DPO $M_2$ & 81.3 & 88.6 & 31.6 & 38.8 & ~~7.0 & ~~8.7 & ~~5.0 & 12.5\\ %
		\rowcolor{gray!7} ~~\method (ours)+DPO $M_1$ & 80.6 & 87.9 & 30.4& 38.4 & ~~7.4& 11.0 & ~~7.5 & 10.0 \\ %
		\rowcolor{gray!7} ~~\method (ours)+DPO $M_2$ & \textbf{81.9} & \textbf{89.0} & \textbf{31.8} & \textbf{39.8} & ~~\textbf{9.3}& \textbf{12.1} & ~~\textbf{7.5} & \textbf{15.0} \\ %
		\midrule
		\rowcolor{white!7} ~~\method (ours) $M_1$ & 80.1 & 87.8 & 32.2& 36.6 & ~~7.9& ~~9.5 & 15.0 & 22.5 \\ %
		\rowcolor{white!7} ~~\method (ours) $M_2$ & 82.8 & 88.9 & 33.2& 38.4 & ~~9.5& 11.7 & 10.0 & 20.0 \\ %
		\bottomrule
	\end{NiceTabular}
	\caption{DPO training results with the preference pair training data curated by our \method method and baseline methods on GSM8K, MATH, Olympiadbench, and AMC23. For instance, IRPO+DPO represents DPO training with the preference data constructed by IRPO method. The results are zero-shot Pass@1 and Maj@8 accuracy.}
	\label{tab: math_DPO}
\end{table*}

\begin{table*}[ht]
	\small
	\centering
	\setlength{\tabcolsep}{7.0pt}
	\resizebox{\textwidth}{!}{
		\begin{NiceTabular}{l|cc|cc|cc|cc}
			\toprule
			{\bf Metric} & \multicolumn{2}{c}{\bf GSM8K} & \multicolumn{2}{c}{\bf MATH-500} & \multicolumn{2}{c}{\bf Olympiadbench} & \multicolumn{2}{c}{\bf AMC23}\\
			~~Iteration & Pass@1 & Maj@8 & Pass@1 & Maj@8 & Pass@1 & Maj@8 & Pass@1 & Maj@8 \\
			\midrule
			\rowcolor{white!7}
			\rowcolor{white!7} {\small \hspace{-2mm} \emph{Llama3-8B-Instruct} } &  & & &  &  &  & &\\
			\rowcolor{white!7} ~~\method (ours)+DPO $M_1$ & 80.6 & 87.9 & 30.4& 38.4 & ~~7.4& 11.0 & ~~7.5 & 10.0 \\ %
			\rowcolor{white!7} ~~\method (ours)+DPO $M_2$ & 81.9 & 89.0 & 31.8 & 39.8 & ~~9.3& 12.1 & ~~7.5 & 15.0 \\ %
			\rowcolor{gray!7} ~~\method (ours) $M_1$ & 80.1 & 87.8 & 32.2& 36.6 & ~~7.9& ~~9.5 & \textbf{15.0} & 22.5 \\ %
			\rowcolor{gray!7} ~~\method (ours) $M_2$ & \textbf{82.8} & 88.9 & \textbf{33.2}& 38.4 & \textbf{~~9.5}& 11.7 & 10.0 & 20.0 \\ %
			\bottomrule
		\end{NiceTabular}
	}
	\caption{\method Loss training and original DPO Loss training results comparison. The results are zero-shot Pass@1 and Maj@8 accuracy.}
	\label{tab: math_loss}
\end{table*}

\subsection{Comparison Results}\label{subsec: main_rusults} 

The main results are shown in Table~\ref{tab: math_main}, demonstrating that the performance of our \method exceeds baseline methods across multiple seed models on the GSM8K, MATH, Olympiadbench, and AMC23 benchmarks.

Specifically, for the Llama-3-8B-Instruct model, \method achieves significant improvements over ScPO and IRPO. On the GSM8K Pass@1 test, it surpasses ScPO and IRPO by 1.2 and 1.7 points, respectively. Similarly, on the MATH-500 Pass@1 test, it outperforms these baselines by 1.0 and 2.6 points, respectively. The improvements are more pronounced on the OlympiadBench Pass@1 test, with gains of 1.6 and 2.8 points over ScPO and IRPO, respectively. Notably, on the AMC23 Pass@1 test, \method achieves an impressive lead of 7.5 and 15.0 points over ScPO and IRPO, respectively. A similar trend is observed for Mathstral-7B-v0.1, with \method achieving gains of 1.4 and 1.3 points on GSM8K Pass@1, 1.4 and 3.4 points on MATH-500 Pass@1, and 5.0 and 12.5 points on AMC23 Pass@1 over ScPO and IRPO.

For the Qwen-2.5-7B-Instruct model and Qwen-2.5-MATH-7B-Instruct model, the performance gains are relatively smaller. We provide a theoretical analysis based on some literature. \citet{mckenzie2023inverse} propose that LMs may show inverse scaling or worse task performance with increased training data scale. And according to \citet{gan2024towards}, the efficacy of large language models (LLMs) is extensively influenced by both the volume and quality of the training data. Qwen-2.5-7B-Instruct model and Qwen-2.5-MATH-7B-Instruct model utilized iterative fine-tuning of data and was reinforced by a reward model during the post-training phase~\cite{yang2024qwen2}. As a result, the quality of the training dataset we use does not significantly benefit the LLM. Nevertheless, \method still consistently outperforms IRPO and ScPO across all benchmarks, demonstrating a clear advantage over outcome-level methods. For the Qwen-2.5-MATH-7B-Instruct model, while IRPO and ScPO underperform the seed model \(M_0\) on MATH-500, \method continues to demonstrate consistent gains, highlighting its robustness over outcome-level methods.

Table~\ref{tab: math_main} also demonstrated that the performance of \method shows more consistency and robustness over the iteration training, detailed explanations in Appendix~\ref{app: iteration}. Overall, \method consistently outperforms the baselines that rely solely on final results when constructing preference training data on all the benchmarks with Pass@1 and Maj@8 metrics.

\section{Ablation Study and Analysis} \label{sec: analysis}

\subsection{Effect of Preference Data}\label{subsec: data}
To isolate the impact of training data quality, we design an experiment where all methods—\method, IRPO~\cite{pang2024iterative}, and ScPO~\cite{prasad2024self}—use the same DPO loss function~\cite{rafailov2024direct}, despite their original loss functions differing as described in Section~\ref{subsec: loss}. This allows us to directly compare the effectiveness of the preference pairs generated by each method.

Table~\ref{tab: math_DPO} shows the performance of Llama-3-8B-Instruct trained with preference pairs curated by \method, IRPO, and ScPO, all optimized using the DPO loss. Here, "IRPO+DPO" denotes training data curated by IRPO with the DPO loss, and similarly for other methods. The results demonstrate that models trained with \method's preference pairs consistently outperform those trained with IRPO or ScPO pairs. Specifically, \method \(M_2\) achieves 1.9 and 2.3 points higher on the OlympiadBench Pass@1 test compared to IRPO \(M_2\) and ScPO \(M_2\), respectively, and 2.5 points higher on the AMC23 Pass@1 test than the best-performing model trained with IRPO or ScPO data. These results highlight the superior quality of \method's preference pairs, further validating its effectiveness in curating training data.

\begin{table*}[htb]
	\small
	\centering
	\setlength{\tabcolsep}{7.0pt}
	\begin{NiceTabular}{l|cc|cc|cc|cc}
		\toprule
		{\bf Metric} & \multicolumn{2}{c}{\bf GSM8K} & \multicolumn{2}{c}{\bf MATH-500} & \multicolumn{2}{c}{\bf Olympiadbench} & \multicolumn{2}{c}{\bf AMC23}\\
		~~Iteration & Pass@1 & Maj@8 & Pass@1 & Maj@8 & Pass@1 & Maj@8 & Pass@1 & Maj@8 \\
		\midrule
		\rowcolor{white!7}
		\rowcolor{white!7} {\small \hspace{-2mm} \emph{Llama3-8B-Instruct} } &  & & &  &  &  & &\\
		\rowcolor{white!7} ~~Seed $M_0$ & 71.3 & 81.6 & 30.8 & 34.2 & ~~8.1 & 10.2 & 10.0 & ~~7.5\\ %
		
		\rowcolor{white!7} ~~DPO $M_1$ & 79.8 & 87.4 & 29.4 & 34.2 & ~~6.8 & 10.8 & ~~5.0 & 10.0\\ %
		\rowcolor{gray!7} ~~\method (ours)+DPO $M_1$& \textbf{80.6} & \textbf{87.9} & \textbf{30.4} & \textbf{38.4} & ~~\textbf{7.4} & \textbf{11.0} & ~~\textbf{7.5} & \textbf{15.0}\\ %
		\rowcolor{white!7} ~~RPO  $M_1$& 79.1 & 86.4 & 29.4 & 35.6 & ~~7.3 & 10.0 & ~~~~0 & 17.5\\ %
		\rowcolor{gray!7} ~~\method (ours)+RPO $M_1$ & \textbf{80.0} & \textbf{87.3} & \textbf{30.6} & \textbf{37.4} & ~~\textbf{7.4} & \textbf{10.7} & ~~\textbf{5.0} & ~~\textbf{22.5} \\ %
		\rowcolor{white!7} ~~IPO $M_1$& 80.6 & 88.0 & 24.4 & 37.6 & ~~8.1 & 11.9 & 10.0 & 10.0\\ %
		\rowcolor{gray!7} ~~\method (ours)+IPO $M_1$& \textbf{81.3} & \textbf{88.1} & \textbf{32.2} & \textbf{38.4} & ~~\textbf{9.9} & \textbf{12.6} & \textbf{15.0} & \textbf{20.0}\\ %
		\rowcolor{white!7} ~~ORPO $M_1$ & 81.6 & 88.1 & 27.0 & 32.8 & ~~8.0 & 10.5 & 10.0 & 12.5 \\ %
		\rowcolor{gray!7} ~~\method (ours)+ORPO $M_1$ & \textbf{81.9} & \textbf{88.2} & \textbf{29.0} & \textbf{36.4} & ~~\textbf{8.6} & \textbf{11.9} & 10.0 & \textbf{25.0} \\ %
		\rowcolor{white!7} ~~TDPO $M_1$ & \textbf{79.8} & 86.5 & 29.8 & 35.0 & ~~7.7 & ~~9.0 & ~~5.0 & 12.5 \\ %
		\rowcolor{gray!7} ~~\method (ours)+TDPO $M_1$ & 79.7 & \textbf{87.1} & \textbf{30.4} & \textbf{36.2} & ~~\textbf{8.4} & ~~\textbf{9.8} & ~~5.0 & \textbf{25.0} \\ %
		\bottomrule
	\end{NiceTabular}
	\caption{Performance of applying \method framework to construct training data with different DPO variants on GSM8K, MATH, Olympiadbench, and AMC23.  The results are zero-shot Pass@1 and Maj@8 accuracy.}
	\label{tab: math_RL}
\end{table*}

\subsection{Effect of Loss}
\begin{figure}[t]
	\centering
	\includegraphics[width=\linewidth]{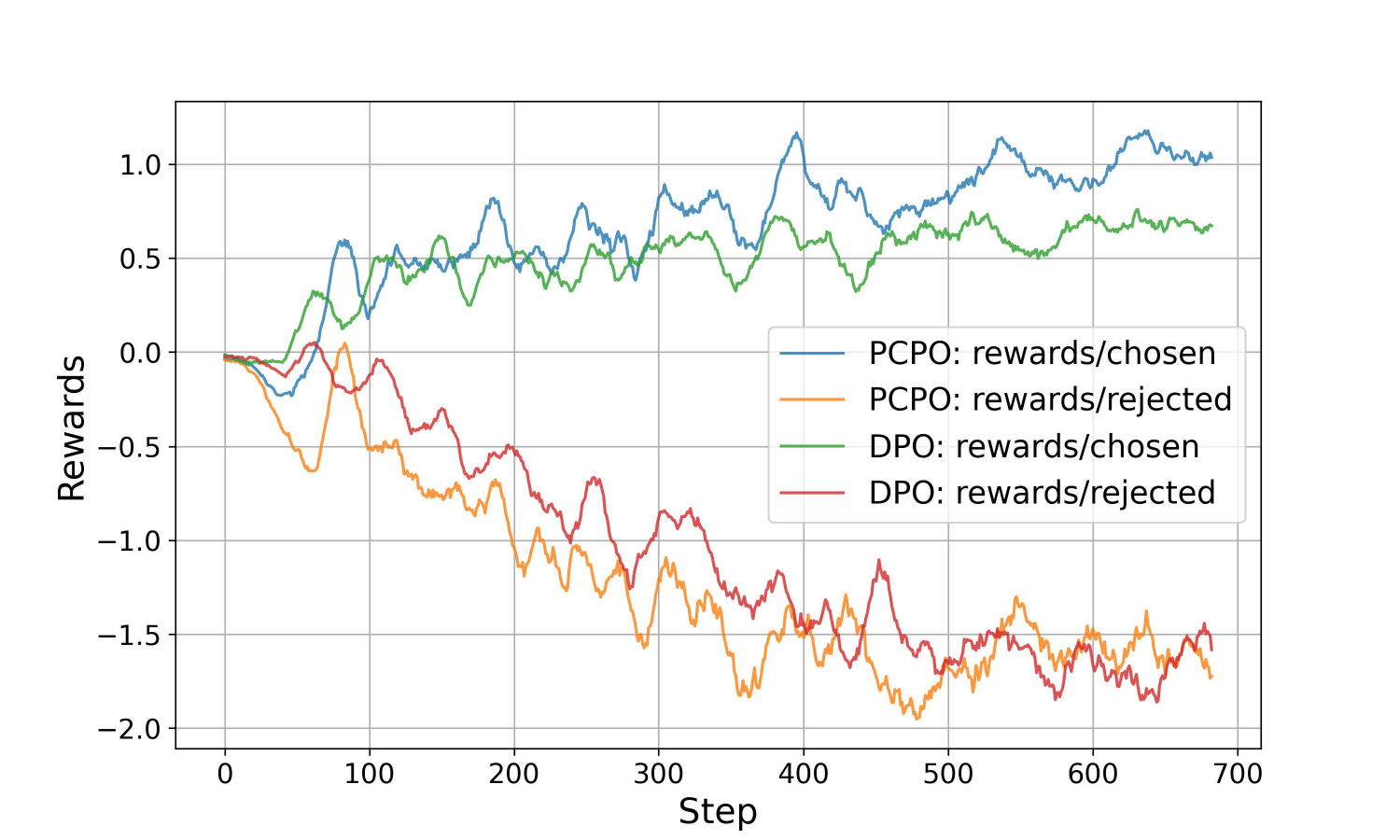}
	\caption{\textbf{Rewards of \method and DPO.} 
		The chosen and rejected responses reward comparison of \method and DPO training on the same preference pairs.
	}
	\label{fig: rewards}
	\vspace{-0.5cm}
\end{figure}

Table~\ref{tab: math_loss} demonstrates that the model trained with the \method Loss, as described in Section~\ref{subsec: loss}, outperforms the model trained with the original DPO Loss on the same \method curated preference pairs.  Figure~\ref{fig: rewards} shows the chosen and rejected responses reward comparison of \method and DPO training on the same preference pairs. The reward, denoted as $r=\beta \log \frac{\pi _{\theta}\left( y|x \right)}{\pi _{ref}\left( y|x \right)}$, reflects the preference intensity of the current strategy model $\pi _{\theta}$ for generating a specific response $\textbf{y}$ relative to the reference model $\pi_{ref}$~\cite{stiennon2020learning,rafailov2024direct}. Notably, the chosen reward for \method Loss exhibits a more pronounced increase, indicating more efficient learning from preference data due to an improved gradient update strategy.
These results underscore that the \method Loss enables more effective preference training compared to the original DPO Loss.

\subsection{Generalizability}\label{subsec: verstaile}
Section~\ref{subsec: data} shows that the preference training data curated by our \method framework is of higher quality. To further validate its versatility, we apply our framework to enhance several DPO variants: RPO~\cite{pang2024iterative} (the single-iteration version of IRPO), IPO~\cite{azar2024general} (designed to prevent overfitting), ORPO~\cite{hong2024orpo} (reference-free alignment), and TDPO~\cite{zeng2024token} (token-level alignment). As shown in Table~\ref{tab: math_RL}, \method+RPO, \method+IPO, \method+ORPO, and \method+TDPO consistently outperform their original counterparts across nearly all benchmarks. These results highlight the effectiveness and broad applicability of our framework in improving diverse preference alignment methods.

\subsection{Case Study}\label{subsec: case}

\begin{figure*}[ht]
	\centering
	\includegraphics[width=\linewidth]{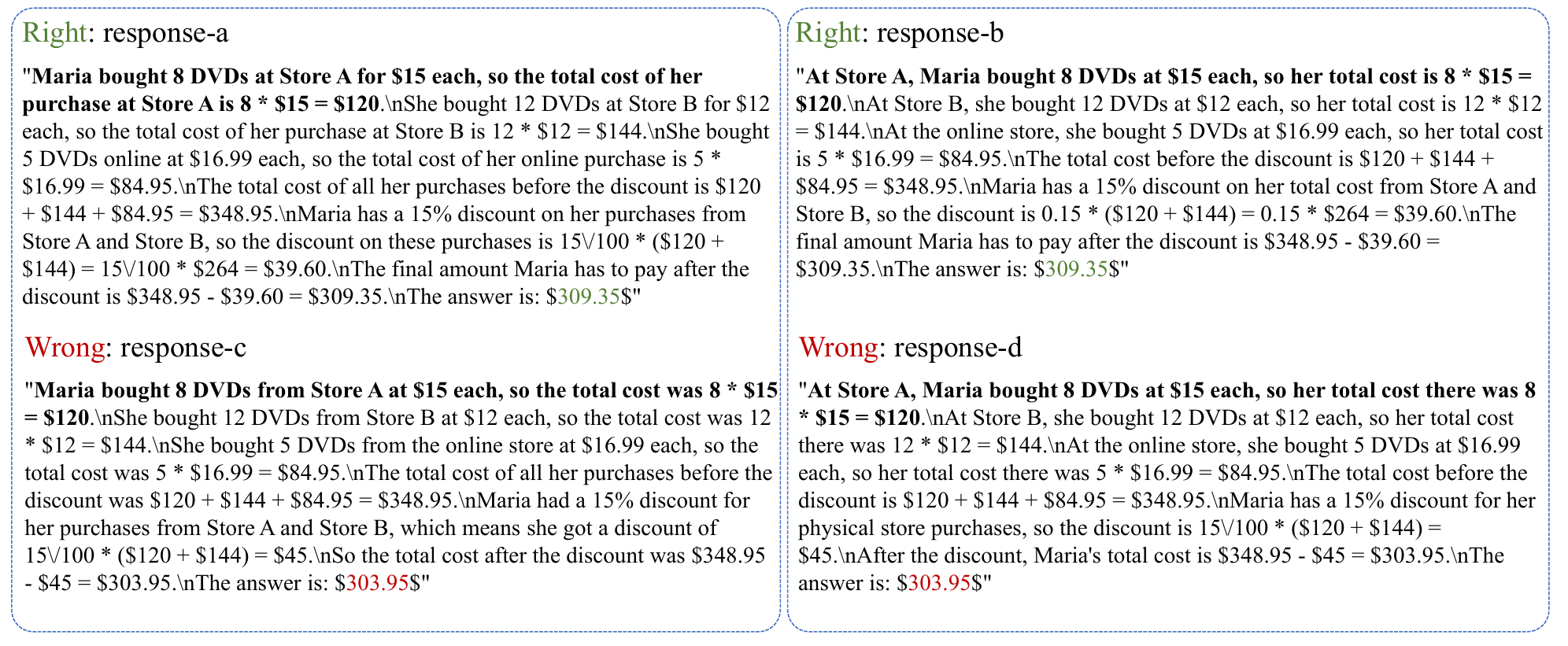}
	\caption{\textbf{A few right and wrong responses from the same prompt.} 
		The four responses can be divided into two groups, where each has a similar response pattern.
	}
	\label{fig: case}
	\vspace{-0.5cm}
\end{figure*}

We have already presented the efficiency and versatility of the preference training pairs curated with our \method framework in Section~\ref{subsec: data}, \ref{subsec: verstaile}, and we will quantitatively analyze it through some cases in this section. Figure~\ref{fig: case} shows four responses from the Llama3-8B-Instruct with the same prompt generated in Section~\ref{subsec: candidate}. These four responses exemplify the response generation process, showcasing both correct and incorrect answers, as well as various answer patterns. In this case, response-a and response-b have the right answer, while response-c and response-d have the wrong answer. Moreover, it can be easily seen from the \textbf{bold} sentence that these four responses have two answer patterns, and response-a and response-c are of one pattern while response-b and response-d are of another pattern. However, because they have no difference in their final answer, outcome-only methods are not able to distinguish them, so it's totally random for these methods to construct preference pairs from them. Our \method can easily identify different answer patterns in the token-level and put the responses with the nearest pattern in a preference pair.
\begin{table}[t]
	\small
	\centering
	\begin{NiceTabular}{l|cc|cc}
		\toprule
		Pairs & a \& c & b \& c & a \& d & b \& d\\
		\midrule
		\rowcolor{white!7} $\mathbf{s_{weighted}}$ & \textbf{0.791} & 0.525 & 0.559 & \textbf{0.793}\\ %
		\bottomrule
	\end{NiceTabular}
	\captionof{table}{The weighted score $s_{weighted}$ of the four responses in Figure~\ref{fig: case}.}
	\label{tab: s_w}
\end{table}

In this case, the weighted scores in Equation~\eqref{eq: score} of these pairs are shown in Table~\ref{tab: s_w}, thus \method is able to select response-a and response-c as a preference pair and response-b and response-d another.
From this case study, we can conclude that our \method can select preference pairs with the highest token probability consistency, which the existing outcome-level methods can not do.

\subsection{Analysis of training consumption}\label{subsec: consumption}

We conducted a statistical analysis of the quantitative comparisons with baseline methods. The entire training process can be divided into three parts: response generation, preference pair construction, and training. The costs of the generating process and training process are nearly identical across all methods, with the main difference arising from the preference pair construction step. We calculated the average computational consumption (converted to 8*A800 GPU Hours) of training seed models for one iteration, as shown in table~\ref{tab: consumption}. 

\begin{table*}[t]
	\centering
	\begin{NiceTabular}{l|c|c|c|c}
		\toprule
		Process & Generate Responses & Construct Preference Pairs & Train & Total\\
		\midrule
		\rowcolor{white!7} IRPO & 3.2 & N/A & 4.5 & 7.7\\ %
		\rowcolor{white!7} ScPO & 3.2 & N/A & 4.5 & 7.7\\ %
		\rowcolor{white!7} PCPO & 3.2 & 1.2 & 4.5 & 8.9\\ %
		\bottomrule
	\end{NiceTabular}
	\captionof{table}{The average computational consumption among PCPO and baseline methods.}
	\label{tab: consumption}
\end{table*}

Although our method requires approximately 15\% more GPU hours in the training process due to the need for token probability calculations, we believe it's worthy the marginal performance gain over the baseline methods.

\section{Related Works} \label{sec:rel}

\paragraph{Preference optimization for math reasoning.}
Reinforcement learning from human feedback (RLHF)~\cite{christiano2017deep} methods such as Direct Optimization Preference (DPO)~\cite{rafailov2024direct} have emerged as a prominent approach for aligning Large Language Models (LLMs) with human preferences~\cite{ouyang2022training,yang2024qwen2}. Recent advancements have introduced specialized variants for mathematical reasoning tasks. For instance, IRPO~\cite{pang2024iterative} selects preference training pairs from generated responses that include a Chain-of-Thought~(CoT) and trains with DPO Loss adding a NLL term. ScPO~\cite{prasad2024self} utilizes a voting function to evaluate the self-consistency~\cite{wang2022self} of responses and trains with a weighted DPO+NLL loss. IPO aims to prevent DPO from overfitting to the preference dataset and ORPO eliminates the need for a reference model. our proposed \method distinguishes itself by explicitly considering the internal logical relationships within preference pairs, offering a unique approach to preference optimization in mathematical reasoning tasks. 

\paragraph{Token-level preference optimization.}
Recent advancements in token-level preference optimization have sought to address the inherent mismatch between sequence-level rewards and the token-level nature of LLM training and generation~\cite{lin2024rho}. For instance, TDPO ~\cite{zeng2024token} introduces a novel framework for aligning LLMs with human preferences at the token level, incorporating forward KL divergence constraints for individual tokens. SparsePO~\cite{christopoulou2024sparsepo} learns automatically during training inherently sparse masks over token-level rewards and KL divergences, highlighting that not all tokens are important in preference optimization. \citet{lin2024critical} illustrated the importance of critical tokens and proposed $c$DPO to automatically recognize and conduct token-level rewards for the critical tokens during the alignment process. The methods above either emphasize or ignore certain tokens when applying preference optimization, while our method \method utilizes token-level probability consistency to select preference pairs before the preference optimization process.

\section{Conclusion}

In this paper, we introduce Probability-Consistent Preference Optimization (PCPO), which provides a quantitative framework for selecting preference pairs by considering both the correctness of the final answer and the internal coherence of the responses. We introduced the concept of token probability consistency and the pair-weighted score to help select resulting preference training pairs. Extensive experiments demonstrate that our method consistently outperforms existing outcome-only criterion approaches (\eg IRPO, ScPO) across a diverse range of LLMs and benchmarks.  This work paves the way for future research aimed at improving reasoning capabilities through more sophisticated preference pair selection methods.

\section*{Limitations}
While our approach demonstrates strong performance in supervised settings, it inherently depends on access to ground-truth final answers to construct reliable preference pairs. Acquiring high-quality labeled data is often resource-intensive, which restricts the scalability of our method to new domains. These limitations require preference optimization frameworks that can function effectively without gold-standard annotations. Addressing these challenges would significantly broaden the applicability of our method to real-world scenarios where labeled data is scarce or unavailable.

Additionally, the process of selecting preference training pairs necessitates generating a substantial number of candidate pairs, which in turn requires producing a larger volume of responses. This increases the computational demands and GPU hours, posing additional resource constraints.

\section*{Ethics Statement}
\subsection*{Privacy Considerations}
In this study, we employed several publicly available datasets, including GSM8K\footnote{\url{https://huggingface.co/datasets/openai/gsm8k}}~\cite{cobbe2021training}, MATH-500\footnote{\url{https://huggingface.co/datasets/HuggingFaceH4/MATH-500}}~\cite{hendrycks2021measuring,lightman2023let}, Olympiadbench\footnote{\url{https://huggingface.co/datasets/realtreetune/olympiadbench}}~\cite{he2024olympiadbench}, AMC23\footnote{\url{https://github.com/QwenLM/Qwen2.5-Math/tree/main/evaluation/data/amc23}}~\cite{AMC2023}, and Numina-math\footnote{\url{https://huggingface.co/datasets/AI-MO/NuminaMath-CoT}}~\cite{li2024numinamath}. These datasets are distributed under permissive licenses.

For model training, we utilized Llama-3-8B-Instruct~\cite{dubey2024llama}, Mathstral-7B-v0.1~\cite{jiang2023mistral}, Qwen-2.5-7B-Instruct~\cite{yang2024qwen2}, and Qwen-2.5-Math-7B-Instruct~\cite{yang2024qwen2math}. All these models are licensed under Apache License 2.0 and are available for academic use.

In summary, our use of these datasets and models strictly complies with ethical guidelines for research data usage, upholding the principles of academic integrity and responsible research conduct.

\subsection*{Security considerations}
Security Considerations
In this study, the models were trained using generated mathematical responses, which were carefully curated to ensure they do not contain any malicious or adversarial content. All responses were derived from fixed problem sets, which were explicitly selected to avoid any overlap with potential test datasets. This approach mitigates the risk of data leakage and ensures that the training process remains secure and unbiased. By adhering to these practices, we maintain the integrity of the training data and prevent any unintended exposure of sensitive or proprietary information.

\section*{Acknowledgement}
This study was supported in part by the Centre for Perceptual and Interactive Intelligence (CPII) Ltd., a CUHK-led InnoCentre under the InnoHK initiative of the Innovation and Technology Commission of the Hong Kong SAR Government, and in part by NSFC-RGC Project N\_CUHK498/24. Hongsheng Li is a PI of CPII under the InnoHK.

\bibliography{references}

\clearpage\appendix\section*{Appendix}

\section{Levenshtein Distance}\label{app: leven}

\begin{figure}[ht]
	\centering
	\includegraphics[width=\linewidth]{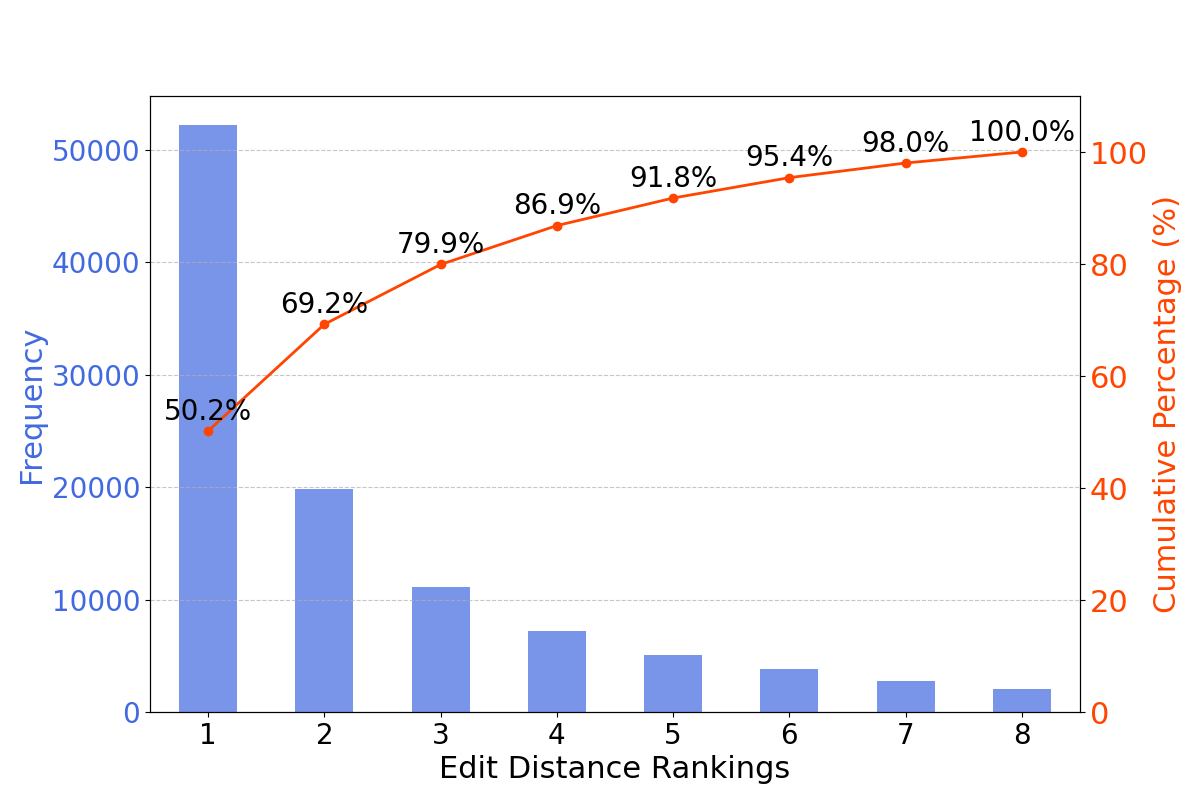}
	\caption{Frequency Distribution and Cumulative Percentage Pareto Chart of Edit Distance Rankings. 
	}
	\label{fig: leven}
	\vspace{-0.5cm}
\end{figure}

In this experiment, we established an edit distance threshold of 8, corresponding to a total of 16 responses per prompt. For each rejected response, we retained the chosen responses with the 8 smallest edit distances. In cases where the number of chosen responses was fewer than 8, all available responses were preserved. Figure~\ref{fig: leven} displays the frequency distribution (bars) and cumulative percentage (line) of edit distance rankings (1–8, from min to max) for the final selected preference pairs.  Rank 1 shows the highest frequency (50.2\%), followed by rank 2 (20\%), with frequencies declining sharply for ranks 5–8. The cumulative percentage reaches 95.4\% by rank 5 and 100\% at rank 8, indicating minimal contributions from higher ranks.
Key Insights:
\begin{itemize}
	\item Pareto Dominance: Ranks 1–5 (95.4\% cumulative) dominate outcomes, aligning with the Pareto principle. 
	\item Central Tendency: Rank 1 alone captures 50.2\%, highlighting strong local consistency.
	\item Low Dispersion: Ranks 6–8 contribute negligibly (<4.6\%), confirming high data concentration.
\end{itemize}

The results mean we can set an edit distance threshold of 5 to filter candidate pairs with more than 90 percent of resulting preference pairs within.
This analysis supports algorithm optimization by prioritizing top-ranked edit distances for candidate pair filtering.

\section{Matching Function}\label{app: match}

\begin{algorithm}[h]
	\small
	\caption{Match Function}
	\label{alg: Match Function}
	\begin{algorithmic}[1]
		\Require $\mathbf{c} = [c_1, \dots, c_m]$, $\mathbf{r} = [r_1, \dots, r_n]$.
		\Ensure Masks $\mathbf{M}_c$, $\mathbf{M}_r$, index mapping $\mathcal{I}$.
		
		\State $\text{matcher} \gets \text{SequenceMatcher}(\text{None}, \mathbf{c}, \mathbf{r})$
		\State $\mathbf{M}_c \gets [\text{False}] \times m$, $\mathbf{M}_r \gets [\text{False}] \times n$
		\State $\mathcal{I} \gets \emptyset$
		
		\For{$(\text{tag}, i_1, i_2, j_1, j_2) \in \text{matcher}$}
		\If{$\text{tag} = \text{equal}$ and $(i_2 - i_1) \geq 1$}
		\For{$(\text{ci}, \text{rj}) \in \text{zip}(\text{range}(i_1, i_2), \text{range}(j_1, j_2))$}
		\State $\mathbf{M}_c[\text{ci}] \gets \text{True}$
		\State $\mathbf{M}_r[\text{rj}] \gets \text{True}$
		\State $\mathcal{I} \gets \mathcal{I} \cup \{(\text{ci}, \text{rj})\}$
		\EndFor
		\EndIf
		\EndFor
		\State \Return $(\mathbf{M}_c, \mathbf{M}_r, \mathcal{I})$
	\end{algorithmic}
\end{algorithm}

Algorithm~\ref{alg: Match Function} shows the match function pseudocode. Let \( \mathbf{c} = [c_1, c_2, \dots, c_m] \) and \( \mathbf{r} = [r_1, r_2, \dots, r_n] \) represent the token sequences of the chosen and rejected responses, respectively. The function identifies the longest common subsequence (LCS) of tokens between \( \mathbf{c} \) and \( \mathbf{r} \). For each aligned subsequence of length at least 1, it generates binary masks \( \mathbf{M}_c \in \{0, 1\}^m \) and \( \mathbf{M}_r \in \{0, 1\}^n \), Additionally, the function outputs an index mapping $\mathcal{I}$, which records the positions of aligned tokens in \( \mathbf{c} \) and \( \mathbf{r} \).

In summary, the function can be compactly represented as
\[
(\mathbf{M}_c, \mathbf{M}_r, \mathcal{I}) = \text{Match}(\mathbf{c}, \mathbf{r})
\]
where \( \text{Match} \) is the sequence matching operation that identifies common tokens and generates the corresponding masks and index mapping.

\begin{figure*}[ht]
	\centering
	\includegraphics[width=\linewidth]{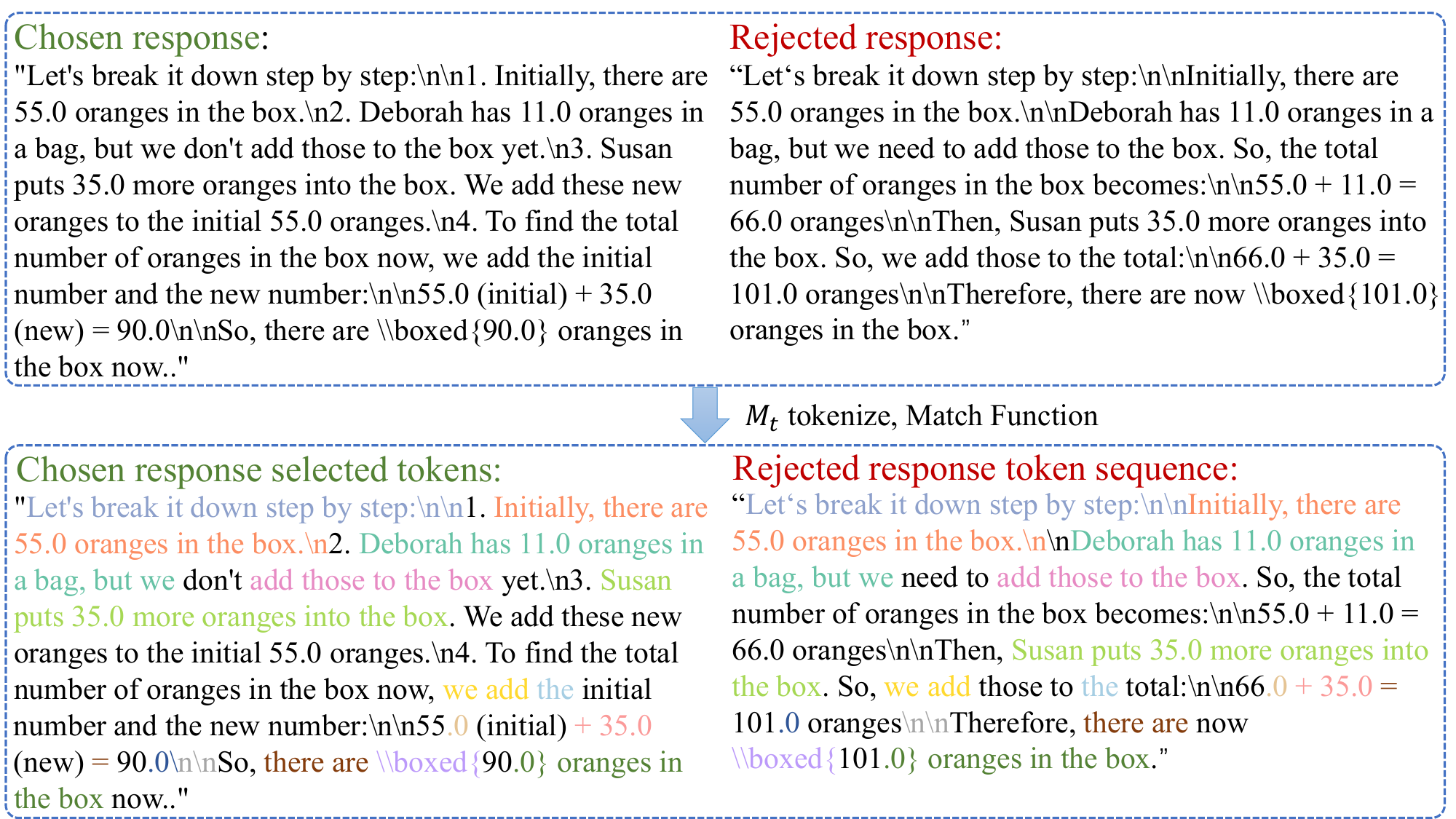}
	\caption{\textbf{The Match Function pipeline.} For a given pair of chosen and rejected responses, we first utilize the current iteration model $M_t$ to tokenize them and then use the algorithm~\ref{alg: Match Function} to get the longest common token subsequences, as highlighted in different colors.
	}
	\label{fig: match}
	\vspace{-0.5cm}
\end{figure*}

Figure~\ref{fig: match} illustrates the visualization for applying the Match function to align token sequences between chosen and rejected responses. As outlined in Section~\ref{subsec: consistency}, we first tokenize the responses using the current iteration model \(M_t\). Next, the Match function \(\mathcal{M}\) generates common token masks for the sequences in a sequential manner. The masked tokens are highlighted in different colors, with index mappings indicating their positions in each sequence. We obtain the final matched tokens by extracting these tokens.

\section{Evaluation details.}\label{app: eval}
We use the standard automatic evaluation scripts following Qwen-Math~\citep{yang2024qwen2math}. The automatic evaluation pipeline mainly contains three steps: response generation, answer parsing, and comparison. First, the pipeline employs model $M_t$ to generate responses for each problem in the test set with a CoT prompt~\cite{kojima2022large} (\eg "Please reason step by step, and put your final answer within boxed\{\}") Second, the pipeline will extract the final answer from the response using regular expressions and fix the format of the answer such as removing extra brackets and modify the representation of fractions \etc Finally, the pipeline compares the extracted answer with the ground truth using an exact match criterion. This criterion requires that the answers satisfy one of the following conditions: (1) numerical equality, where both answers can be converted to floats and are equal, or (2) symbolic equality, where both answers can be converted to \textsc{sympy}\footnote{\url{https://github.com/sympy/sympy}} expressions and are equal. Through this pipeline, we can maximize the consistency and accuracy of the test results.

\section{Iterations}\label{app: iteration}
Table~\ref{tab: math_main} presents the model performance evolvement over the seed model $M_0$, $M_1$ and $M_2$. In a nutshell, \method performs better along the iterations over baseline methods IRPO, ScPO while achieving better absolute scores. For instance, on the Llama3-8B-Instruct,\method Pass@1 on the GSM8K test evolves from $M_1$ 80.1\% to $M_2$ 82.8\%, and the Olympiadbench test evolves from $M_1$ 7.9\% to $M_2$ 9.5\%, surpassing each iteration of the IRPO and ScPO method. Results on the Mathstral-7B-v0.1 show a similar trend. Although the iteration gains of all methods on the Qwen2.5-7B-Instruct and the Qwen2.5-MATH-7B-Instruct is less stable owning to the reason we explained in Section~\ref{subsec: main_rusults}, our \method still performs a more consistent performance. The ScPO method almost saturates on the Qwen2.5-7B-Instruct through iterations, with only a small gain on the AMC23 Pass@1 test, and the IRPO method drops on the GSM8k, MATH-500 and AMC23 Pass@1 test from $M_1$ to $M_2$. 
In all, the performance of \method shows more consistency and robustness over the iteration training, confirming the effectiveness of our method.

\section{Prompts}
\label{app:prompts}
Prompt templates\footnote{The prompt template was from \url{https://github.com/QwenLM/Qwen2.5-Math}} for generating responses are shown below:

\begin{tcolorbox}[colback=blue!5!white, colframe=blue!80!black, title=Response Generation Template]
	\textbf{User:}
	\begin{verbatim}
		Please reason step by step, and put 
		your final answer within \\boxed{{}}.
		
		{{ question }}
	\end{verbatim}
	\textbf{Assistant:}
\end{tcolorbox}

\end{document}